\PassOptionsToPackage{pagebackref,breaklinks,colorlinks}{hyperref}
\documentclass[10pt,twocolumn,letterpaper]{article}

\usepackage[pagenumbers]{wacv} 
\usepackage{amsmath,amssymb,amsfonts}
\usepackage{algorithmic}
\usepackage{graphicx}
\usepackage{textcomp}
\usepackage{todonotes}
\usepackage{makecell}
\usepackage{multirow, booktabs}
\usepackage{comment, xcolor}
\usepackage{orcidlink}
\DeclareRobustCommand\onedot{\futurelet\@let@token\@onedot}
\usepackage[capitalize]{cleveref}
\crefname{section}{Sec.}{Secs.}
\Crefname{section}{Section}{Sections}
\Crefname{table}{Table}{Tables}
\crefname{table}{Tab.}{Tabs.}

\def\wacvPaperID{1517} 
\def\confName{WACV}
\def\confYear{2025}

\newcommand{\authorblock}[1]{\begin{tabular}{@{}c@{}}#1\end{tabular}}

\author{\begin{tabular}{c@{\qquad}c}
  \authorblock{Xiaohu Liu\\
SJTU Paris Elite Institute of Technology, \\ Shanghai Jiao Tong University\\
Shanghai, China\\
{\tt\small liuxiaohu@sjtu.edu.cn}
  \\ 
  } &
  \authorblock{Sascha Hornauer\\
Center for Robotics, \\ MINES Paris, PSL University\\
Paris, France\\
{\tt\small sascha.hornauer@minesparis.psl.eu}
  } \\[\bigskipamount]\\[\smallskipamount]
  \authorblock{Fabien Moutarde\\
Center for Robotics, \\ MINES Paris, PSL University\\
Paris, France\\
{\tt\small fabien.moutarde@minesparis.psl.eu}
  } & 
  \authorblock{Jialiang Lu\\
SJTU Paris Elite Institute of Technology, \\ Shanghai Jiao Tong University\\
Shanghai, China\\
{\tt\small jialiang.lu@sjtu.edu.cn}
  }
\end{tabular}}

\begin{document}


\title{AVS-Net: Audio-Visual Scale Net for Self-supervised Monocular Metric Depth Estimation}

\maketitle


\begin{abstract}

Metric depth prediction from monocular videos suffers from bad generalization between datasets and requires supervised depth data for scale-correct training. Self-supervised training using multi-view reconstruction can benefit from large scale natural videos but not provide correct scale, limiting its benefits. Recently, reflecting audible Echoes\footnote{All the modalities, including "Echoes" and others, start with a capital letter.} off objects is investigated for improved depth prediction and was shown to be sufficient to reconstruct objects at scale even without a visual signal. Because Echoes travel at fixed speed they have the potential to resolve ambiguities in object scale and appearance. However, predicting depth end-to-end from sound and vision can not benefit from unsupervised depth prediction approaches, which can process large scale data without sound annotation. In this work we show how Echoes can benefit depth prediction in two ways: When learning metric depth learned from supervised data and as supervisory signal for scale-correct self-supervised training. We show how we can improve the predictions of several state-of-the-art approaches and how the method can scale-correct a self-supervised depth approach. 
\end{abstract}


\section{Introduction}
The objective of monocular depth estimation is to estimate the depth map from a single RGB camera image. This approach offers a cost-effective and conceptually simple alternative to stereo depth estimation \cite{laga2020survey} or Lidar based active methods \cite{park2018high}. 
It can be applied to many real-world scenarios and large-scale datasets with no available depth information. Due to its potential benefits, it has been the subject of extensive investigation
\cite{ming2021deep,bhat2021adabins,bhat2023zoedepth,bhat2022localbins,zhou2017unsupervised,zhang2023lite,zhao2022monovit}.


\begin{figure}[!tbp]
\centerline{\includegraphics[width=1.0\linewidth]{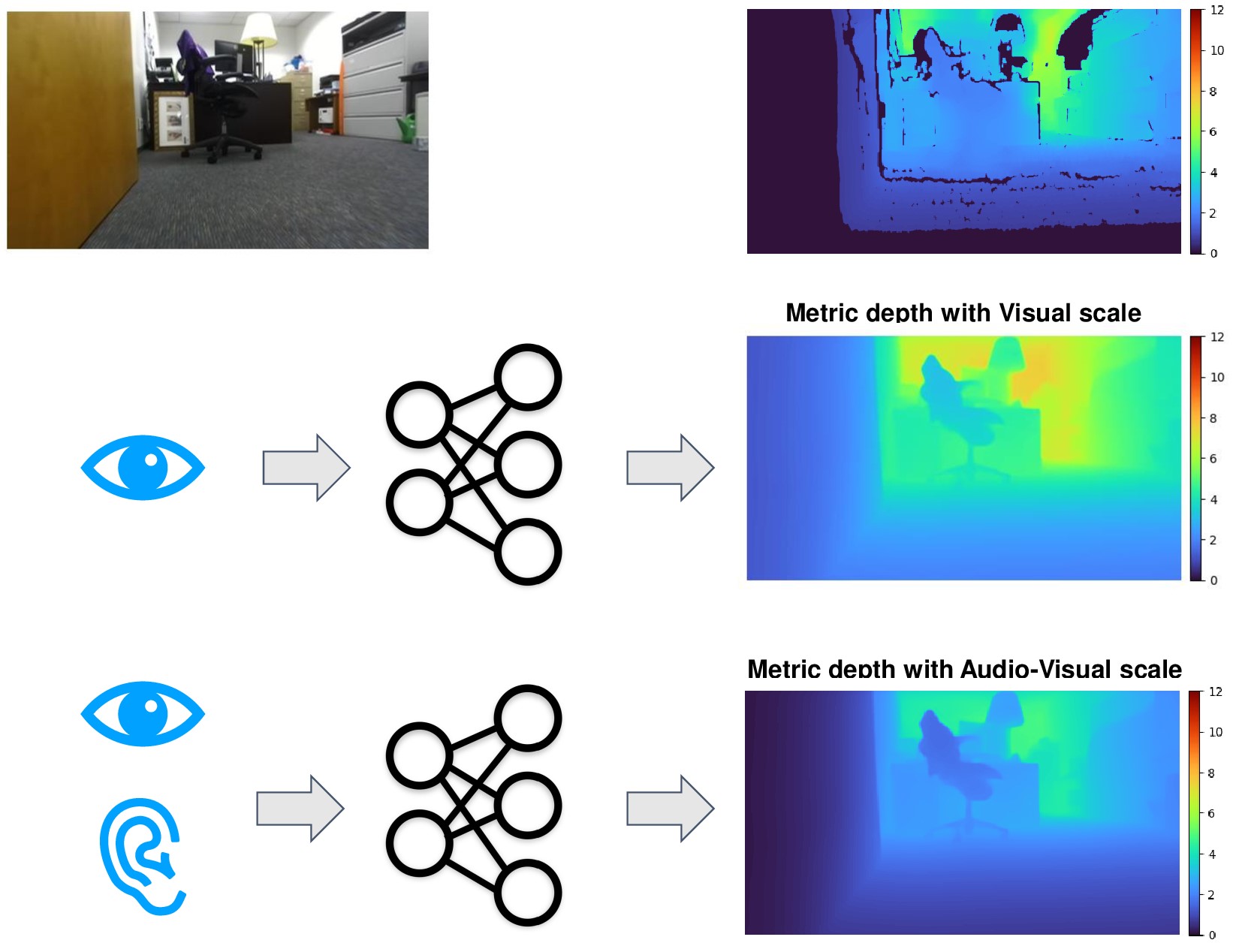}}
\caption{Visual-only scaling can provide erroneous results (middle row). While Echoes, which inherently contains scale information, can help resolve such scale ambiguity (down row).}
\label{fig:intro}
\end{figure}

At the same time, monocular depth estimation remains an ill-posed problem because the same RGB image can correspond to an infinite number of 3D scenes\cite{ming2021deep}. To address these issues, a number of methods have been proposed in recent years. These include the use of surface normal maps\cite{qi2018geonet}, gradient maps\cite{jun2022depth}, or combining depth prediction with auxiliary tasks such as semantic segmentation \cite{zhang2018joint}. For improved accuracy, regression to depth pixel values can be reformulated as a classification problem \cite{fu2018deep,bhat2021adabins,bhat2022localbins,bhat2023zoedepth}. 


The distinction between \textit{metric} \cite{bhat2021adabins,bhat2022localbins,li2024binsformer,bhat2023zoedepth,jun2022depth} and \textit{relative} \cite{godard2019digging,zhao2022monovit,zhang2023lite,yin2018geonet} depth estimation can be made on the basis of the presence or absence of absolute physical scale information (meters), as provided by the predicted depth map. The former returns the actual physical distance but it tends to overfit to a particular dataset and exhibits poor generalisation performance \cite{bhat2023zoedepth}. The latter approach predicts the relative distance between pixels, can be trained across multiple datasets \cite{ranftl2020towards} and exhibits better generalisation performance. However, the absence of physical scale information limits its application in domains such as autonomous driving and robot navigation.

Concurrently, the integration of sound, using Echoes to supervised depth estimation has shown robust improvement of visual depth quality and helped to resolve visual illusions \cite{chen2022visual,vedaldi_visualechoes_2020,parida2021beyond,liupano,brunetto2023audio}. Echoes, reflected from objects in the scene, travel at fixed speeds and can inform the network reliably about true distances. Nevertheless, no investigation has been conducted into the potential of Echoes to bridge the gap between relative and metric depth-based techniques.

To achieve scale-corrections for the self-supervised relative depth estimation method\cite{godard2019digging}, we propose a two-step method to explore the metric information contained in Echoes. First we add audio to a recently proposed metric depth prediction method for improved supervised model training on a audio-visual dataset \cite{bhat2023zoedepth}. Then, scale factors are extracted to provide the missing scale information for relative depth models such as \cite{godard2019digging,zhao2022monovit,zhang2023lite} or correct pre-trained zero-shot models such as \cite{bhat2023zoedepth,jun2022depth,yuan2022neural}.

In \cite{bhat2023zoedepth}, a similar approach has been adopted by first pre-training on 12 datasets with relative depth and then fine-tuning using metric depth. However in our proposed method, the AVS-Net firstly learns the relationship between Echoes and correct metric depth. Then the trained model is used to scale-correct other models, which are trained on datasets with relative depth. In this way, relative depth prediction methods can improve and continuously learn on new audio-visual datasets while our correction-model only re-scales the results informed by Echoes. 



Our main contributions can be summarized as follows: 

\begin{itemize}
\item We propose AVS-Net, a new method using Echoes to provide scale information for self-supervised relative- and zero-shot metric-depth models.
\item Extensive evaluations on the BatVision datasets \cite{brunetto2023audio} show that our proposed Audio-Visual method significantly outperformed the visual-only counterpart by approximately 30\% and 22\% in terms of $\delta_{1}$ accuracy for the first-stage pseudo-dense\footnote{The rationale behind the introduction of the \textit{pseudo-dense} metric depth is to facilitate differentiation from the final metric depth, i.e. depth map combining metric factors and relative depth} metric depth and the final scale-corrected metric depth map respectively. This demonstrates the effectiveness of Echoes in providing valuable geometry scale information.
\item The proposed method can be easily transferred to other relative- or zero-shot metric-depth models. 
\end{itemize}

\section{Related work}

\subsection{Monocular metric depth estimation}

Monocular Depth estimation can be roughly divided into two categories: \textit{Relative depth estimation} \cite{zhou2017unsupervised,godard2019digging,zhao2022monovit,zhang2023lite, ranftl2020towards} predicts relative spatial relationships regardless of the absolute distances. Relative depth-based methods can be trained across different datasets while maintaining good generalization ability \cite{ranftl2020towards}. However, the loss of scale information constrains their real-world application scenarios. \textit{Metric depth estimation} \cite{bhat2023zoedepth,jun2022depth,li2024radarcam, bhat2021adabins,yuan2022neural,bhat2022localbins,piccinelli2024unidepth} estimates the depth map with physically correct distances but tends to overfit on individual datasets and suffers from distribution shift \cite{bhat2023zoedepth}. 

A recent trend in metric depth estimation is the use of metric bins (i.e. discretization) within depth intervals \cite{bhat2021adabins, bhat2022localbins, li2024binsformer, bhat2023zoedepth}.
Based on the observation that reformulating metric depth regression as a classification task can improve the performance \cite{fu2018deep}, Bhat et al., \cite{bhat2021adabins} proposed Adabins which estimate adaptive depth bin centers according to the scene of the input image. The final estimated metric depth is obtained by linear combination of the bin centers to overcome discontinuities. LocalBins \cite{bhat2022localbins} further extends Adabins by employing multi-scale decoder features for local histogram estimation instead of a global one. The similar idea of employing multi-scale features has also been tested in BinsFormer \cite{li2024binsformer}, which uses a transformer decoder to predict bins. ZoeDepth \cite{bhat2023zoedepth} has proposed a novel approach that combines relative depth models and metric depth bins. It involves pre-training the relative depth model on multiple datasets via Midas \cite{ranftl2020towards} to exploit its generalisation capability. The lightweight metric bins module is then trained on a metric dataset to provide domain-specific scale parameters. During inference, input images are fed to a corresponding metric bins head by the classification network which mitigates overfitting to specific datasets. 

Another approach is to divide metric depth estimation into relative depth estimation and metric coefficients predictions \cite{jun2022depth, li2024radarcam}. Scale factors are derived from feature vectors \cite{jun2022depth}, or from heterogeneous modalities \cite{li2024radarcam} (e.g. sparse radar point cloud).

This work shares a similar objective with that presented by Li et al. \cite{li2024radarcam} in that the aim is to combine extracted scale factors from heterogeneous modalities and pre-trained relative depth models. However, our work differs in that we explore Echoes, which inherently contains metric scale information. We get this benefit while sound is a low-cost and simple to use modality with sensors often already integrated in video recording devices.

\subsection{Audio-Visual Learning}
Improving depth prediction using sound is an increasingly active field. Early works show proof-of-concept results using audible Echoes like sonar in simulation and on real world data
\cite{christensen_batvision_2020, vedaldi_visualechoes_2020}. They are useful when encountering transparent or reflecting surfaces, leading to ambiguous interpretations from vision alone \cite{ye_3d_2015, kim_3d_2017, liupano}. Simulated and real-world audio-visual datasets and simulators are being created and support this domain \cite{dai_binaural_2023, brunetto2023audio}.
Sounds arrive from 360° and even around corners which has been exploited for improved panoramic depth prediction and even reconstruction outside the field of view \cite{purushwalkam_audio-visual_2021, zhu_beyond_2024, liupano}. The Echoes-based principle is the basis of our work to predict scale-correct depth images.

\cite{chen_sound_2023} jointly learn sound direction and camera rotation. By exploiting consistency in camera movement and the perceived direction of arrival of sounds they can define a self-supervised learning objective. We show in this work how to benefit from self-supervised training but correct the relative depth to correct metric scale.

\begin{figure}[htbp]
\centerline{\includegraphics[width=1.0\linewidth]{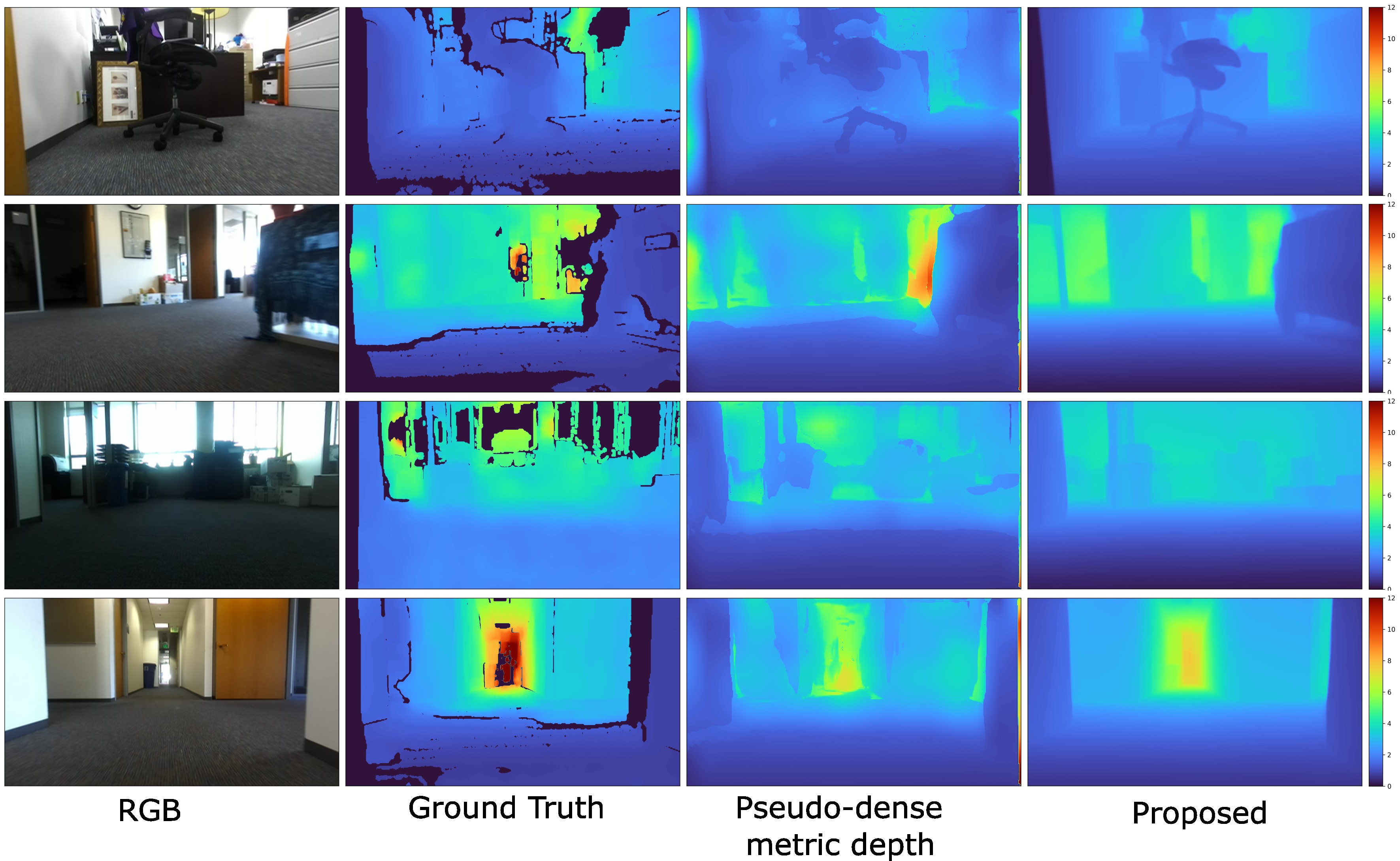}}
\caption{Comparisons between pseudo-dense metric depth and final scale-corrected depth (proposed). Though the former provides globally correct scale information, the depth quality degrade because of over-fitting to low quality Ground Truth. While the later benefit from both correct scale and superior depth quality}
\label{fig:intro_effect}
\end{figure}

\section{Method}

\begin{figure*}[htbp]
\centerline{\includegraphics[width=0.9\linewidth]{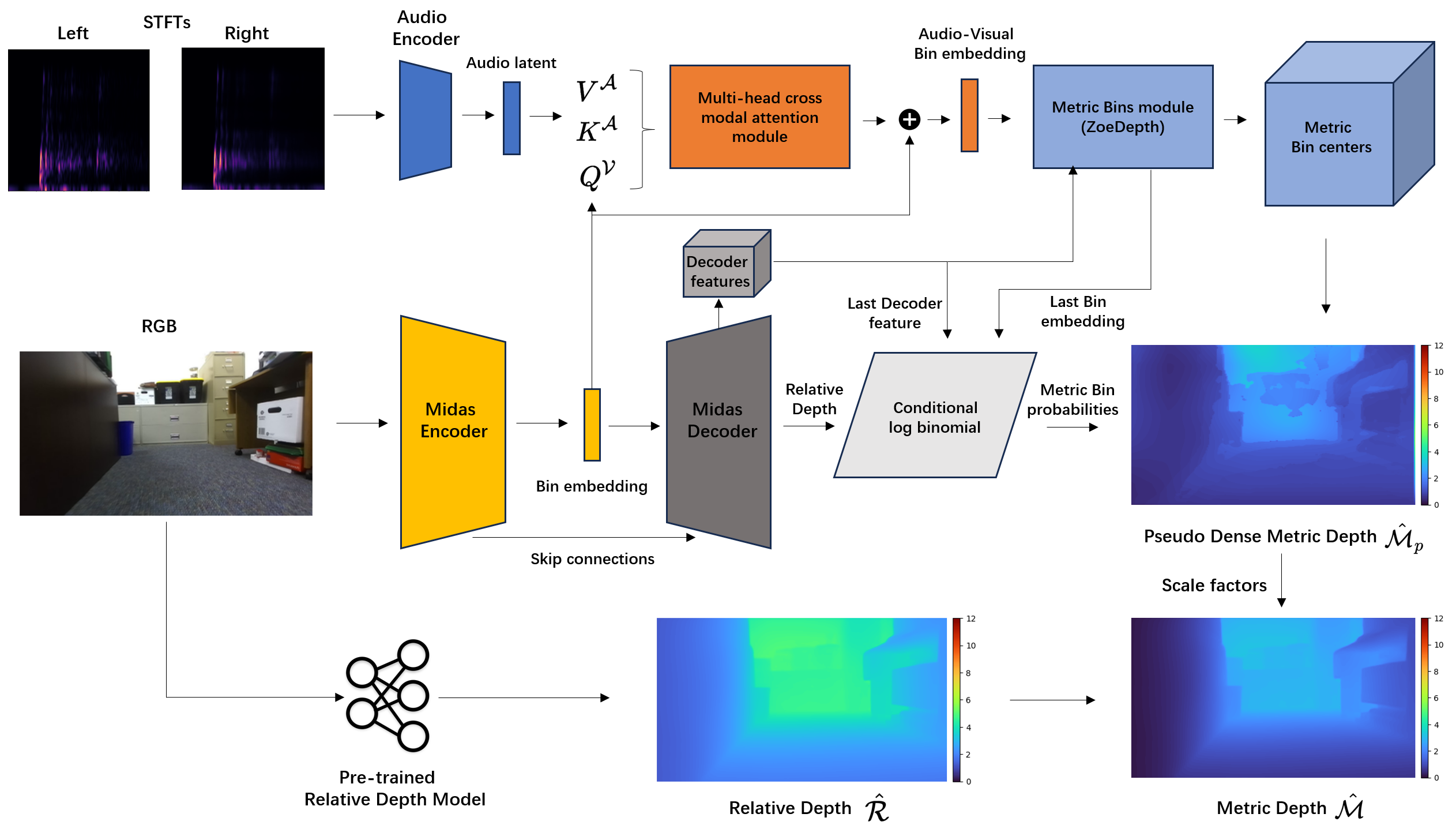}}
\caption{Illustration of the proposed AVS-Net (upper two rows), which takes binaural STFT and RGB as input. The Audio-Visual latent vectors are fused by a multi-head cross attention module, to estimate metric bin centers and pseudo-dense metric depth. Scale factors are extracted from the later to combine with a relative depth map from pre-trained relative depth model and form the final metric depth.}
\label{fig:main-pipeline}
\end{figure*}

We propose Audio-Visual ScaleNet (AVS-Net), which uses Echoes for better self-supervised metric depth estimation. The intuition of our approach is to decompose metric depth estimation into relative depth estimation and metric factor estimation in order to combine the generalisation ability of relative depth models and to use Echoes, which inherently contain scale information, for better scale parameter estimation.

Our approach is shown in Fig.\ref{fig:main-pipeline}. Compared to ZoeDepth \cite{bhat2023zoedepth}, the two phases in our model are permuted: in the first phase, Audio-Visual fusion of STFT Short-Time Fourier Transform) representations of Echoes is performed to better estimate metric bin centers and obtain enhanced pseudo-dense metric maps. In the second phase, the enhanced pseudo-dense metric maps are used to obtain scale factors to combine with other pre-trained unsupervised relative depth models or zero-shot models for metric depth estimation. AVS-Net uses a pre-trained Midas \cite{ranftl2020towards} encoder to obtain visual feature vectors. The Echoes feature vector is obtained by adding an additional Echoes embedding branch to encode ego-centric Echoes corresponding to the input RGB image. Audio-Visual fusion is performed by a cross-modal attention module \cite{liupano,gulati2020conformer}. These fused vectors are then passed through the \textit{seed bin regressor} \cite{bhat2023zoedepth} to obtain the initial bin centres, which will be incrementally adjusted by the attraction layers proposed in ZoeDepth, and finally the metric bins are linearly combined by a log-binomial to obtain the pseudo-dense metric depth map.

The pseudo dense metric depth map obtained from supervised training in the first stage is overfitted for a particular dataset. We show an effect from the Batvision V1 dataset, whose artifacts in the ground truth show as discontinous depth in the prediction and therefore degrades the quality of the relative depth (e.g., Fig. \ref{fig:intro_effect}). However, we suggest the scale information can still be used and so we extract scale parameters from the pseudo dense metric depth map and combine them with the pre-trained relative depth model to achieve the final metric depth prediction. The two stages are described more in details below.


\subsection{Audio-Visual ScaleNet} The goal of the first stage is to improve the pseudo-dense metric depth prediction using ego-centric Echoes with RGB images. 

\textbf{Visual embedding} We encode the input image with a pre-trained Midas \cite{ranftl2020towards} encoder to obtain bin embeddings $C$ from the bottleneck. We use them as the visual vector $f^{\mathcal{V}}$ to be fused.

\textbf{Audio embedding} We transform the stereo Echoes sampled at 44100 Hz into a two-channel magnitude STFT of size 2$\times$F$\times$T with Short-Time Fourier Transform. Note that 2 stands for left and right and not for phase information which we dismiss, we have F frequency bins and T overlapped time windows. The 2D time-frequency representation is resized to the same resolution as input RGB image and fed to a ResNet-18 encoder resulting in audio embedding $f^{\mathcal{A}}$. While many ways exist to process STFT's, we are inspired from \cite{majumder2022few} which processes room impulse responses that have similarities with our chirp.

\textbf{Audio-Visual fusion} The majority of existing audiovisual fusion methods employ either channel dimension concatenation for feature superposition and channel-wise  dimensionality reduction \cite{vedaldi_visualechoes_2020}, or bilinear transformations and attention masks for adaptive attention to a particular modality \cite{parida2021beyond}. Cross-modal attention \cite{chen2022visual,liupano} has been found to offer better performance than audio-visual feature concatenation. Consequently, we use multi-head cross modal attention to extract useful information in the Echoes STFT queried by the visual modality. We compute a visual query $Q^{\mathcal{V}}$ via visual feature embedding $f^{\mathcal{V}}$, and auditory key $K^{\mathcal{A}}$ and value $V^{\mathcal{A}}$ are obtained from the auditory embedding $f^{\mathcal{A}}$ via the cross-modal attention module:

\begin{equation}
Att\left( f^{\mathcal{V}},f^{\mathcal{A}}\right)=\operatorname{softmax}\left(\frac{Q^{\mathcal{V}} (K^{\mathcal{A}})^T}{\sqrt{s}}\right) V^{\mathcal{A}}
\label{eq:crossmodalEqu1}
\end{equation}

where $Q^{\mathcal{V}}= W_{Q}^{\mathcal{V}}f^{\mathcal{V}}, K^{\mathcal{A}}= W_{K}^{\mathcal{A}}f^{\mathcal{A}}, V^{\mathcal{A}}= W_{V}^{\mathcal{A}}f^{\mathcal{A}}$ and $s$ denotes the scaling factor. The skip connection with the original bin embedding $f^{\mathcal{V}}$ will be performed to get final Audio-Visual bin embedding:

\begin{equation}
L^{\mathcal{A},\mathcal{V}}=Att\left( f^{\mathcal{V}},f^{\mathcal{A}}\right) + f^{\mathcal{V}}
\label{eq:crossmodalEqu2}
\end{equation}

\textbf{Pseudo dense metric depth estimation}
The $L^{\mathcal{A},\mathcal{V}}$ is then fed into the \textit{seed bin regressor} and \textit{seed projector} \cite{bhat2023zoedepth} to produce the initial metric bin centers and initial bin embedding. These are subsequently adjusted through attractor layers described in \cite{bhat2023zoedepth} with the help of other high-resolution decoder features. The metric bin probabilities volume will be obtained from the last high-resolution decoder feature, in addition to the Midas \cite{ranftl2020towards} relative depth. This will be used with adjusted bin centers through a log-binomial to produce the pseudo-dense metric depth, denoted as $\hat{\mathcal{M}}_{p}$.

\subsection{Self-supervised Metric Depth with Echoes} The aforementioned pseudo-dense metric depth could be used as a metric depth output. However, effects such as a poor quality ground truth depth map in one dataset (e.g. the upper right corner in Fig. \ref{fig:intro}), can decrease the performance of an overfit model on other datasets. 
Nevertheless, scale information can still be correct in the presence of artifacts. We therefore extract scale parameters from $\hat{\mathcal{M}}_{p}$ and combine them with either a self-supervised relative depth model 
to provide missing metric factors, or a pre-trained zero-shot model 
for metric scale corrections. This allows to take advantage of the generalisation ability of the relative depth (or depth from zero-shot models) combined with Echoes to provide scale information.


\begin{figure}[htbp]
\centerline{\includegraphics[width=1.0\linewidth]{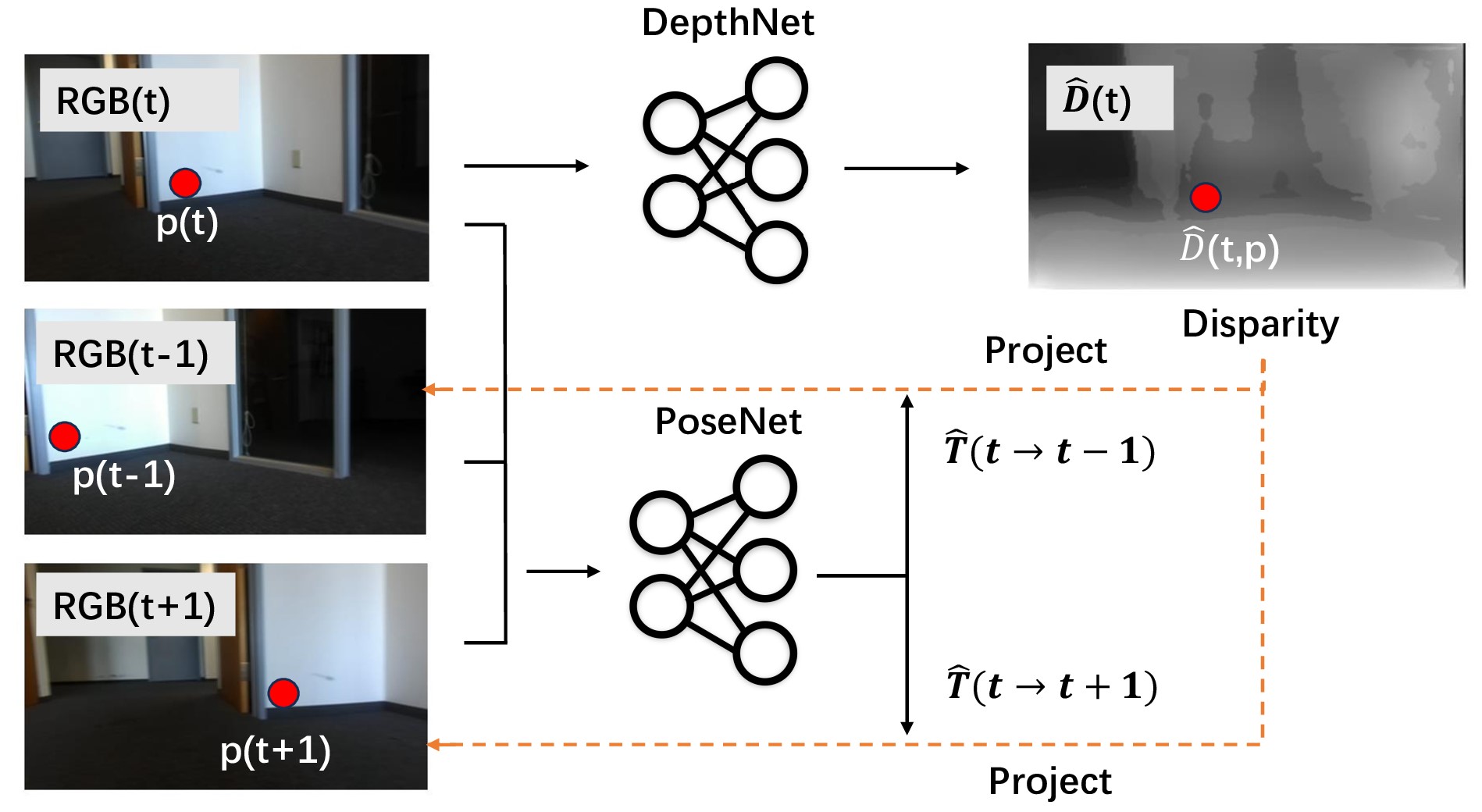}}
\caption{Illustration of the self-supervised relative depth estimation method, inspired by multi-view approaches such as \cite{godard2019digging,zhou2017unsupervised}}
\label{fig:self-sup-training}
\end{figure}

\subsubsection{Self-supervised relative depth estimation}
\label{subsec:Self-supervisedrelativedepthestimation}
A prominent category of relative depth estimation methods is the joint estimation of depth (or disparity) map and camera pose using only continuous video sequences \cite{zhou2017unsupervised,godard2019digging,zhao2022monovit,zhang2023lite}. This approach eliminates the necessity for supervised methods for ground truth depth maps, offering enhanced generalisation capabilities. However, the training process only requires the predicted disparity map and camera motion to match, without involving the physical scale information. Therefore, the obtained depth map and camera motion vectors are relative quantities. 

We showcase scale-correction of a self-supervised depth prediction method in this work. Training is depicted in Fig. \ref{fig:self-sup-training}. A video frame from a sequence at the current time step $RGB(t)$ is input into \textit{DepthNet} as target view, with the objective of estimating the disparity map at the current time step $\hat{D(t)}$. Concurrently, the adjacent frames $RGB(t-1)$ and $RGB(t+1)$ are employed as source views, together with the current frame, as input into $PoseNet$ to predict the 6$DoF$ camera motion vectors of the adjacent frames, which are represented as $\hat{T}(t\xrightarrow{} t-1)$ and $\hat{T}(t\xrightarrow{}t+1)$, respectively. The predicted disparity map is then projected onto the point cloud and combined with the camera motion variables in order to determine the position of the pixels in the source views at the current time step. This process synthesises the predicted target views from the source views. The joint training of $DepthNet$ and $PoseNet$ is ultimately achieved through the minimisation of the photometric reprojection error \cite{zhou2017unsupervised}. Though joint training, for inference, DepthNet is capable of predicting the relative depth associated with the input RGB image on its own, denoted as $\hat{\mathcal{R}}$.

\subsubsection{Audio-Visual Scaling} In the second stage, scale factors are extracted from the pseudo-dense metric depth $\hat{\mathcal{M}}_{p}$ and combined with the relative depth $\hat{\mathcal{R}}$ to obtain the final metric depth $\hat{\mathcal{M}}$.

As suitable scale factors, mean and standard deviation are tested in \cite{jun2022depth, li2024radarcam}. However, they are sensitive to outliers in both $\hat{\mathcal{M}}_{p}$ and $\hat{\mathcal{R}}$. \textit{Median} values are less influenced by outliers and frequently employed in self-supervised joint learning-based methods \cite{zhou2017unsupervised,godard2019digging,zhao2022monovit,zhang2023lite}. In our evaluation we compare and find clear improved performance of median scaling over mean-std \footnote{results for mean-std scaling can be found in supplementary materials}. It is applied as follows:

\begin{equation}
\hat{\mathcal{M}} = \hat{\mathcal{R}}*\frac{\mathtt{MEDIAN}(\hat{\mathcal{M}}_{p})}{\mathtt{MEDIAN}(\hat{\mathcal{R}})}
\label{eq:median_scaling}
\end{equation}

\subsection{Model training}
The supervised training for the AVS-Net and self-supervised training for relative depth models are performed as follows: 

\subsubsection{AVS-Net} 
The AVS-Net is trained fully supervised with the objective of minimising the discrepancy between $\hat{\mathcal{M}_{p}}$ and ground truth label $\mathcal{M}$ with Scale-Invariant (SI) loss. Introduced in \cite{eigen2014depth}, SI loss aims to remove the global scale from the loss element, i.e. the loss does not change when the estimated depth is multiplied by a constant. Following  \cite{bhat2023zoedepth,bhat2021adabins,bhat2022localbins,lee2019big}, we use a scaled version of SI Loss for the training of AVS-Net:

\begin{equation}
\mathcal{L}_{\text {SI }}=\alpha \sqrt{\frac{1}{T} \sum_i g_i^2-\frac{\lambda}{T^2}\left(\sum_i g_i\right)^2}
\label{eq:SIloss}
\end{equation}

where $g_i=\log \hat{\mathcal{M}}_{p}({i})-\log {{\mathcal{M}}}(i)$, $T$ is the valid pixel number for 
${\mathcal{M}}$. $\lambda$ controls the weight of error variance, while $\alpha$ ease the convergence by properly scale the loss values \cite{lee2019big}. Same as \cite{bhat2021adabins,bhat2023zoedepth}, we choose $\alpha = 10, \lambda = 0.85$ for our experiments. The resolution of 256$\times$512 is employed for both RGB and $\mathcal{M}$.

\subsubsection{Self-supervised depth models} For the self-supervised training of relative depth models, following \cite{godard2019digging}, we employed a multi-scale joint loss function composed of $L1$ loss, $SSIM$ (e.g. Structured Similarity
Index Measure \cite{wang2004image}) loss and disparity smoothness loss $\mathcal{L}_{\text {smothness}}$ \cite{godard2019digging} between estimated and ground truth disparity maps $\hat{D}$ and ${D}$ for the joint training of $PoseNet$ and $DepthNet$, with per-pixel minimum photometric loss \cite{godard2019digging} defined as:

\begin{equation}
\mathcal{L}_{\text {pe}}^{i}= \min_{t^{\prime}}(\beta\mathcal{L}_{\text {1}}^{i}(t,t^{\prime}) + \gamma\mathcal{L}_{\text {SSIM}}^{i}(t,t^{\prime}))
\label{eq:peloss}
\end{equation}

where $\mathcal{L}_{\text {1}}^{i}(t,t^{\prime})$ denotes $L1$ loss between synthesized frame in time step $t^{\prime} \in \{t-1,t+1\}$ for scale index $i\in\{1,\frac{1}{2},\frac{1}{4},\frac{1}{8}\}$\footnote{The scale index will be changed to $i\in\{1,\frac{1}{2},\frac{1}{4}\}$ for experiments based on Lite-Mono\cite{zhang2023lite}}. The joint loss function is as follows:

\begin{equation}
\mathcal{L}_{\text {joint}}= \frac{1}{N}\sum_i^N (\mathcal{L}_{\text {pe}}^{i} + \lambda\mathcal{L}_{\text {smoothness}}^{i})
\label{eq:jointloss}
\end{equation}

Following \cite{godard2019digging,zhao2022monovit,zhang2023lite}, the weight coefficients are chosen as  $\beta=0.15,\gamma=0.85,\lambda=1e-3$. Each frame is resized to the resolution $192\times640$ to align with previous works \cite{godard2019digging,zhao2022monovit,zhang2023lite}. More training details are in the supplementary.

\begin{table*}[!htb]
\caption{Quantitative results of AVS-Net on test set of BV2 and BV1, with RGB-Echoes denotes proposed Audio-Visual method with additional Echoes branch, Only-RGB denotes the baseline without additional Echoes branch (ZoeDepth). The two methods were trained on the train set of BV2 and tested separately on the test set of BV2 (the second and third rows) and the test set of BV1 (last two rows) to show the zero-shot transfer capability of the proposed method.}

\centering
\small{
\begin{tabular}{ccccccccc}
\hline Dataset & Models & $\mathbf{Abs\  rel}$ $\downarrow$ & $\mathbf{sq\  rel}$ $\downarrow$ & $\mathbf{R M S E} \downarrow$ & $\mathbf{R M S E}(\mathbf{l o g}) \downarrow$  & $\boldsymbol{\delta}_{\mathbf{1}} \uparrow$ & $\boldsymbol{\delta}_{\mathbf{2}} \uparrow$ & $\boldsymbol{\delta}_{\mathbf{3}} \uparrow$
\\
\hline \hline \multirow{3}{*}{ \makecell{BV2}} 
& BITD \cite{parida2021beyond} & 0.323 & - & 2.286 & - & 0.647 &  0.834 & 0.901 \\
& AVS-Net (Only-RGB)  & 0.189 & 0.233 & 0.696 & 0.253 & 0.745 &  0.938 & 0.978 \\
& AVS-Net (RGB-Echoes) & \textbf{0.170} & \textbf{0.210} & \textbf{0.678} & \textbf{0.245} & \textbf{0.776} & \textbf{0.942} & \textbf{0.976}\\
\hline
\multirow{2}{*}{ \makecell{BV1}} 
 & AVS-Net (Only-RGB)$^{\ast}$  & 0.315 & 0.518& 1.149& 0.479& 0.414& 0.680& 0.842   \\
 & AVS-Net (RGB-Echoes)$^{\ast}$ & \textbf{0.277} & \textbf{0.450}& \textbf{1.006}& \textbf{0.398}& \textbf{0.537}& \textbf{0.792}& \textbf{0.906} \\

\hline
\multicolumn{6}{l}{
$^{\ast}$Zero-shot inference results with models trained on the BV2 training set and tested on the BV1 test set.}
\end{tabular}
}
\label{tab:ScaleNet}
\end{table*}

\begin{table*}[!htb]
\caption{Quantitative results with median scaling on the test set of BV1. Monodepth2, MonoVit and LiteMono are trained self-supervised on train set of BV1, ZoeDepth, Jun et al., and NeWCRFs are pre-trained zero-shot metric depth models} 

\centering
\small{
\begin{tabular}{ccccccccc}
\hline Models & Scaling  Method & $\mathbf{Abs\  rel}$ $\downarrow$ & $\mathbf{sq\  rel}$ $\downarrow$ & $\mathbf{R M S E} \downarrow$ & $\mathbf{R M S E}(\mathbf{l o g}) \downarrow$ & $\boldsymbol{\delta}_{\mathbf{1}} \uparrow$ & $\boldsymbol{\delta}_{\mathbf{2}} \uparrow$ & $\boldsymbol{\delta}_{\mathbf{3}} \uparrow$ \\
\hline \hline \multirow{3}{*}{ \makecell{Monodepth2}\cite{godard2019digging}} &  No  & 0.649 & 0.931& 1.656& 1.168& 0.008& 0.023& 0.065  	 \\
& RGB & 0.312 & 0.354& 0.975& 0.498& 0.397& 0.677& 0.827
	 \\ 
& RGB-Echoes 
& \textbf{0.289} & \textbf{0.318}& \textbf{0.898}& \textbf{0.433}& \textbf{0.490}& \textbf{0.753}& \textbf{0.874}
 	 \\
\cmidrule(r){2-9}
\multirow{3}{*}{ \makecell{MonoVit}\cite{zhao2022monovit}}  & No  & 0.568 & 0.757& 1.490& 0.998& 0.017& 0.072& 0.283 	 \\
& RGB  & 0.320 & 0.386& 1.002& 0.531& 0.393& 0.673& 0.814
\\
& RGB-Echoes & \textbf{0.297} & \textbf{0.355}& \textbf{0.931}& \textbf{0.465}& \textbf{0.484}& \textbf{0.746}& \textbf{0.860}
\\
 \cmidrule(r){2-9}
\multirow{3}{*}{ \makecell{LiteMono}\cite{zhang2023lite}}  & No   & 0.782 & 1.284& 1.933& 1.626& 0.000& 0.000& 0.001\\
& RGB & 0.284 & 0.313& 0.951& 0.458& 0.436& 0.719& 0.852
\\
& RGB-Echoes & \textbf{0.261} & \textbf{\underline{0.270}} & \textbf{\underline{0.875}} & \textbf{0.393}& \textbf{\underline{0.525}} & \textbf{\underline{0.785}} & \textbf{0.898}
	 \\
\cmidrule(r){1-9}
\multirow{3}{*}{ \makecell{ZoeDepth}\cite{bhat2023zoedepth}}  & No(zero-shot)  & 1.000 & 1.630& 1.558& 0.705& 0.107& 0.258& 0.486 \\ 
& RGB & 0.285 & 0.394& 1.181& 0.455& 0.420& 0.677& 0.829
\\ 
& RGB-Echoes  & \textbf{\underline{0.250}} & \textbf{0.301}& \textbf{1.042}& \textbf{\underline{0.377}} & \textbf{0.512}& \textbf{0.780}& \textbf{\underline{0.901}}
\\ 

\cmidrule(r){2-9}

\multirow{3}{*}{ \makecell{Jun et al.,}\cite{jun2022depth}}  & No(zero-shot)  & 0.549 & 0.597& \textbf{1.083}& 0.483& 0.241& 0.532& 0.828 \\ 

& RGB & 0.303 & 0.437& 1.258& 0.493& 0.387& 0.647& 0.807
\\ 

& RGB-Echoes & \textbf{0.260} & \textbf{0.342}& 1.129& \textbf{0.409}& \textbf{0.495}& \textbf{0.755}& \textbf{0.880}
\\

\cmidrule(r){2-9}

\multirow{3}{*}{ \makecell{NeWCRFs \cite{yuan2022neural}}}  & No(zero-shot)  & 1.240 & 2.485& 1.874& 0.813& 0.094& 0.208& 0.388 \\ 

& RGB & 0.309 & 0.453& 1.287& 0.505& 0.386& 0.630& 0.793
 \\ 

& RGB-Echoes & \textbf{0.276} & \textbf{0.361}& \textbf{1.156}& \textbf{0.425}& \textbf{0.463}& \textbf{0.737}& \textbf{0.869}
\\ 
\hline
\hline 

\end{tabular}
}
\label{tab:Median_scaling_results}
\end{table*}

\begin{figure*}[htbp]
\centerline{\includegraphics[width=\linewidth]{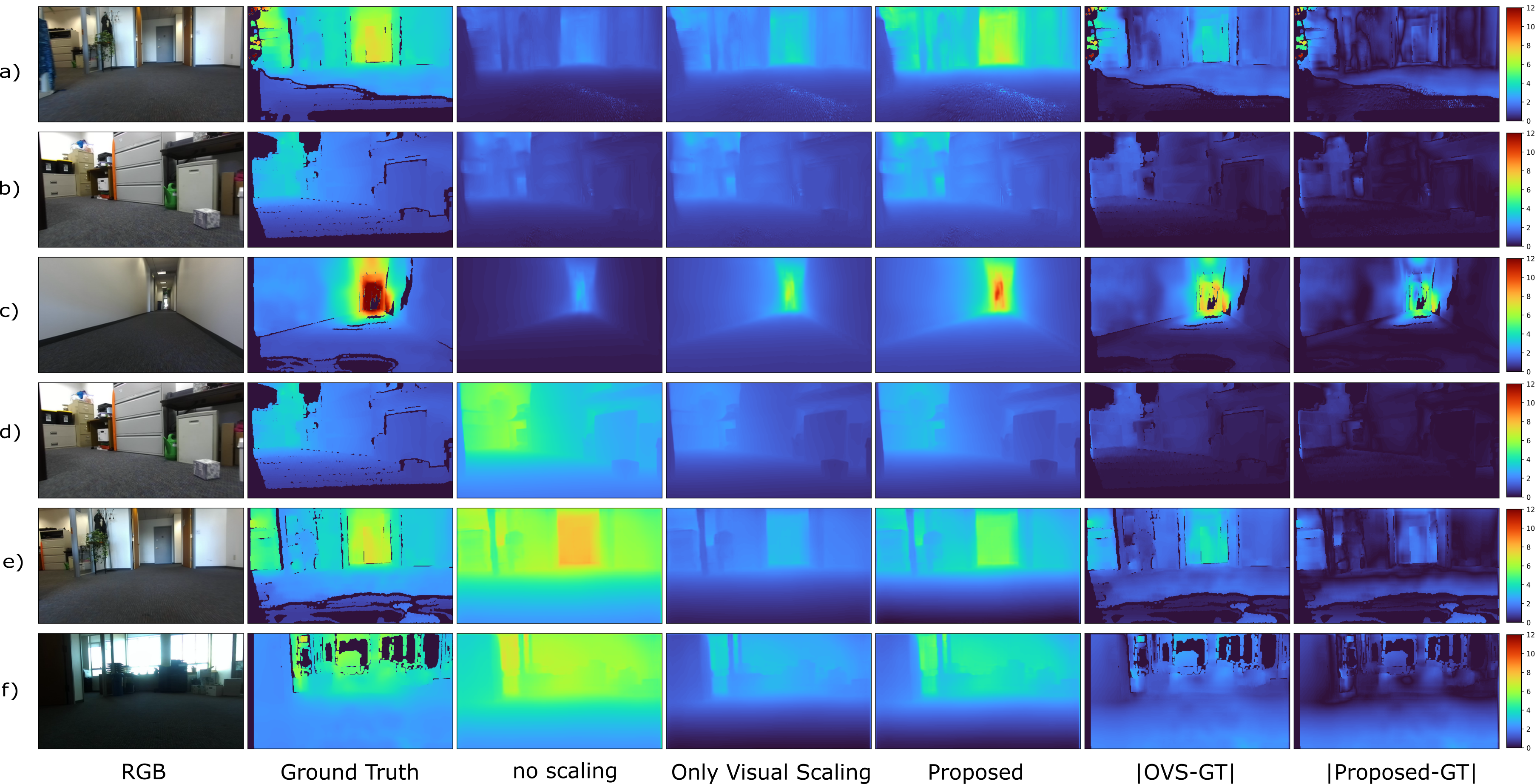}}
\caption{Qualitative examples, with Proposed denotes RGB-Echoes scaling. from a) to f) are based on Monodepth2, MonoVit, LiteMono, ZoeDepth, NeWCRFs, Jun et al., respectively. From left to right: input RGB image, Ground truth depth, estimated depth without scaling, estimated depth with Visual only scaling (AVS-Net(only-RGB)), with RGB-Echoes scaling (AVS-Net(RGB-Echoes)), absolute difference between Only Visual Scaling (OVS) and ground truth, and between proposed and ground truth}
\label{fig:qualitative_results1}
\end{figure*}

\section{Experiments}

\subsection{Dataset}


We evaluate the performance of the proposed approach on the recently released BatVision dataset \cite{brunetto2023audio}, which contains two subsets, named $BV1$ and $BV2$, recorded at different places. Both contain RGB-D pairs synchronized in time with recorded binaural Echoes signals from emitted \textit{Chirps} ranging from 20Hz to 20kHz. These datasets have very different average depth, being recorded in offices and in an historic university. We showcase the generalization between datasets by training the AVS-Net only on the training split of $BV2$, and the relative depth model on the training split of $BV1$. Both test splits will only used for evaluations.

For the supervised training of AVS-Net, both RGB-D and Echoes are needed for training, validation and test. We follow the train-val-test split of $BV2$ \cite{brunetto2023audio}, resulting in 1911 data instances for training, 625 for validation and 584 for test. For each binaural audio signal, we follow original BatVision settings \cite{brunetto2023audio} to generate the STFT 
and resize the STFT magnitude to the resolution of the RGB image before feeding it to the Audio Encoder. For the self-supervised training of relative depth models, only consecutive videos frames are needed. For the evaluation of the proposed pipeline, both RGB-D and Echoes are needed in the test set. While the original $BV1$ set contains 52,220 instances, we re-organise the frames to form consequent frame triples (Ref. \ref{subsec:Self-supervisedrelativedepthestimation}) and eliminate the intermediate change of scenes based on the official train-val-test split. We choose time interval equals to 20 frames for noticeable while reasonable camera movements. Finally 39165 RGB frame triples are eligible for train, 7378 for validation. We keep the left 4960 RGB-D-Echoes data instance for final test, with same settings for STFTs used as for $BV2$.

\subsection{Evaluation metrics}
We resize all predicted depth maps to the same resolution as the original ground truth map ($720\times1280$) and only use valid pixels\footnote{valide mask are calculated from pixels with ground truth depth non zero and strictly smaller than maximum, 12m for BV1 and 30m for BV2.}. For evaluation metrics, we adopted the commonly used Abs Rel, Sq Rel, RMSE, RMSE log, and three threshold accuracies $\delta_{1} < 1.25$,
$\delta_{2} < 1.25^{2}$, $\delta_{3} < 1.25^{3}$.

\begin{figure*}[htbp]
\centerline{\includegraphics[width=\linewidth]{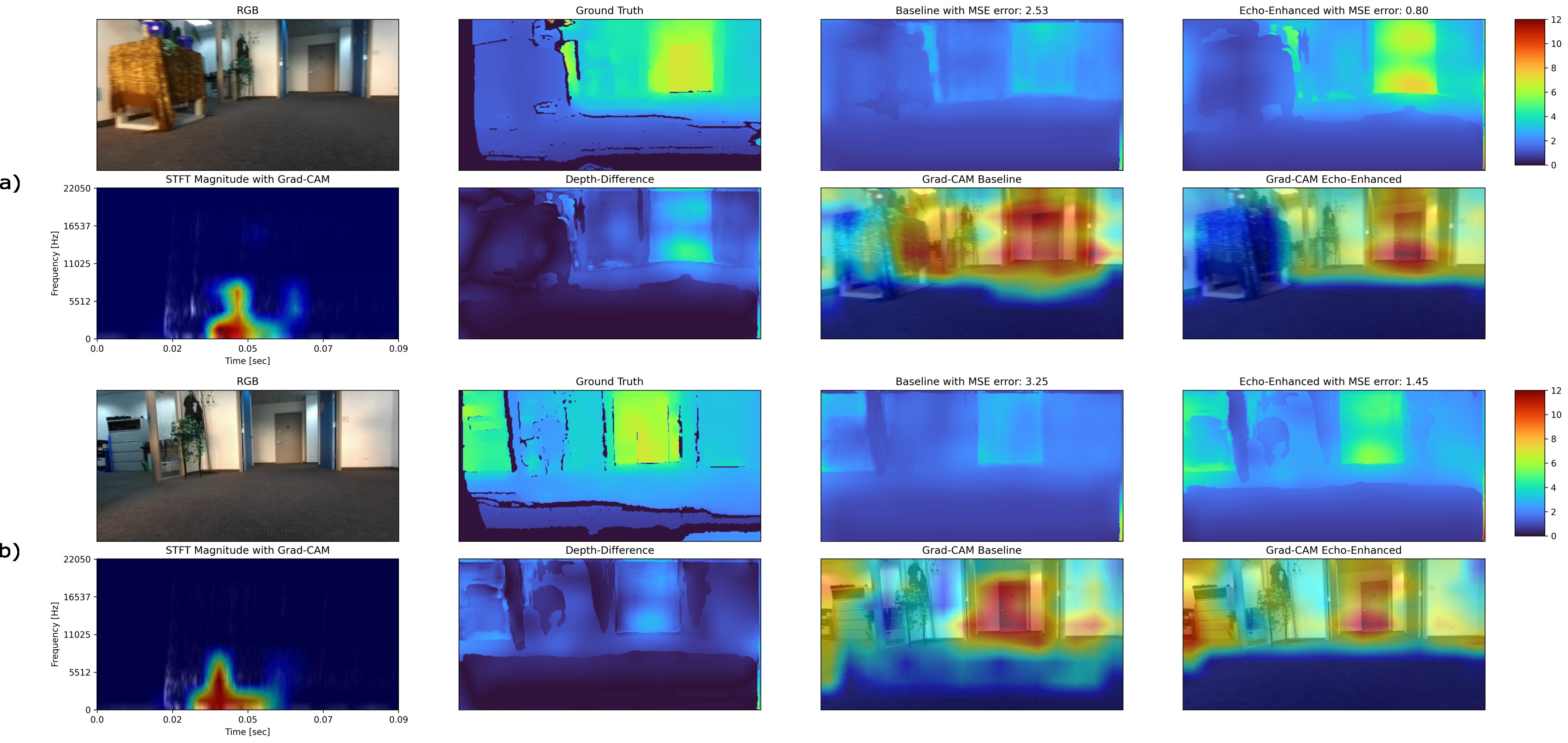}}
\caption{Grad-CAM \cite{selvaraju2017grad} for STFT and RGB for Pseudo-dense metric depth (AVS-Net), with estimated metric depth with (Echoes-enhanced) and without (Baseline) Audio branch. First line from left to right: input RGB image, Ground Truth depth, estimated depth for baseline, estimated depth for Echoes-enhanced. Second line from left to right: Attention map with STFT, difference of two estimated depth maps (abs), attention map on RGB for baseline, attention map on RGB for Echoes-enhanced model.}
\label{fig:grad-cam}
\end{figure*}


\subsection{Results} With our experimental setup we achieve enhanced pseudo metric depth with Echoes and show the effectiveness of AVS-Net for metric factors. We also investigated learned features with Grad-CAM \cite{selvaraju2017grad} and finally discuss limitations of this approach.

\textbf{Enhanced Pseudo Metric Depth with Echoes} We showcase the effectiveness of Echoes in enhancing supervised pseudo metric depth estimation in Tab \ref{tab:ScaleNet}. Though trained only on BV2, we tested on both test set of BV2 and BV1 to illustrate the generalization ability to unseen data. With the help of Echoes, the AVS-Net (RGB-Echoes) surpassed the only visual baseline (Only-RGB) by a margin of about 10\% for both Abs Rel and Sq Rel on BV2. On BV1, the improvement margins are further enlarged to about 12\% for Abs Rel and 13\% for Sq Rel. Especially for $\delta_{1}$ accuracy, the proposed method with Echoes achieved about 23\% improvement compared to its only-visual counterpart, 
demonstrating good generalization. Note that while BV1 and BV2 are published as one dataset, they contain significantly different scenes, depth profiles and sensors used \cite{brunetto2023audio}.

For comparison with other SOTA Audio-Visual methods, we duplicated the results of Parida \textit{et al}.\cite{parida2021beyond} in Tab \ref{tab:ScaleNet}. For the results on BV2, AVS-Net(Only-RGB) already significantly outperformed the current Audio-Visual SOTA by large margins. This could be attributed to the transfer learning and the network architecture. If comparison is made only within Audio-Visual methods, the even larger improvements of AVS-Net(RGB-Echoes) compared with Parida \textit{et al}.\cite{parida2021beyond} 
demonstrate AVS-Net effectively employs the Echoes information for metric depth estimation.





\textbf{Effectiveness of AVS-Net for metric factors} We ablate using Echoes for improved (or corrected) scale by comparing several relative depth models \cite{godard2019digging,zhao2022monovit,zhang2023lite} and pre-trained zero-shot models \cite{bhat2023zoedepth,jun2022depth,yuan2022neural} on BV1, employing different scaling approaches, shown in Tab \ref{tab:Median_scaling_results}. Because relative depth-based models are not designed for evaluation of metric depth, the results are presented for all models and the three different scaling methods for fair comparison. "No" denotes no scaling, RGB scaling means scale parameters are derived from AVS-Net (Only-RGB) and RGB-Echoes scaling denotes the full method for obtaining scale factors from AVS-Net (RGB-Echoes). Our proposed Audio-Visual scaling method surpassed all the baseline method without Echoes, showing good performance providing robust metric scale factors. Compared with visual baselines, improvements of about 10\% in Abs Rel and Sq Rel, about 19\% for $\delta_{1}$ are obtained for relative depth models (first 3 lines), while for zero-shot models (last 3 lines) about 20\% improvements for Sq Rel and $\delta_{1}$ are observed. This shows the utility of Echoes for correcting scale parameters.

Qualitative illustrations in figure \ref{fig:qualitative_results1} show the visual scaling approach offers a reasonable estimation of scale but is unable to accurately predict the distance from obstacles or corridors in the distance. In contrast, audio-visual scaling resolves distance ambiguity through incorporating distance information derived from STFT (cmp. \ref{fig:grad-cam}). This reduces the discrepancy between estimated and actual distances in comparison to the ground truth (last two columns). Please note that the final quality of the metric depth depends also on the relative depth part, thus on different models and training schemes (i.e. Comparing Proposed column in Fig. \ref{fig:qualitative_results1}).

\textbf{Grad-CAM analysis}
We are interested how the AVS-Net uses the STFT input. An expectation would be to see a correlation of network attention over time (areas left or right on the abscissa of the STFT) and depth in the image. Therefore, we calculate the attention map using Grad-CAM \cite{selvaraju2017grad}. In figure \ref{fig:grad-cam} we show plausible attention, linked to areas whose depth is more difficult to asses. In examples (a) and (b), the same physical location is shown. In (b), the robot has advanced compared to (a). The two examples demonstrate improved estimation of distance for distant walls and doors. The attention map for the input STFT for (a) and (b) demonstrate a concentration at specific time-frequency regions.
In (a), the focus of attention is concentrated at a specific point in time, occurring at approximately 0.045 seconds. This corresponds to a distance of approximately 4 meters (at a speed of sound at 340m/s), which is approximately the distance of the blue-green wall in the far distance. In case (b) the attention peak shifts to the left on the time axis, which is a logical consequence of the robot's advancement, as in this case the distance to the far door has decreased, necessitating a focus on smaller time values in the STFT. In addition, the peak of attention shifted by about 1 meter, which seems to correspond to the movement of the robot.


\textbf{Limitations} One of the limitations of this work is that the proposed global scale factor method is unable to resolve the artefacts caused by relative depth models (see (a) in Fig. \ref{fig:qualitative_results1}). Instead, it only provides missing or corrects erroneous scale factors. The future direction of this research will involve the implementation of a post-processing network and the incorporation of Echoes into the relative depth model. This will facilitate the removal of visual illusions and artefacts \cite{liupano, parida2021beyond, vedaldi_visualechoes_2020}.

\section{Conclusion}
In this paper we propose AVS-Net which employs Echoes for enhanced pseudo-dense estimation. Scale factors can be derived from sound and used to furnish missing scale information for relative depth, or to rectify pre-trained zero-shot metric depth models. The experimental results demonstrate that the proposed Audio-Visual method markedly outperforms the visual-only baseline, thereby substantiating the efficacy of Echoes in retrieving scale-correct depth. The plug-and-play nature of the AVS-Net allows for straightforward integration with other relative depth or zero-shot metric depth models, thereby making the method applicable in a wide range of scenarios. 


\bibliography{PaperForReview}
\bibliographystyle{IEEEtran}

\end{document}